# Multi-level Stress Assessment from ECG in a Virtual Reality Environment using Multimodal Fusion

Zeeshan Ahmad, *Member, IEEE,* Suha Rabbani, *Member, IEEE,*
Muhammad Rehman Zafar, *Member, IEEE,* Syem Ishaque, *Member, IEEE,*
Sridhar Krishnan, *Senior Member, IEEE,* and Naimul Khan, *Senior Member, IEEE*

**Abstract**—ECG is an attractive option to assess stress in serious Virtual Reality (VR) applications due to its non-invasive nature. However, the existing Machine Learning (ML) models perform poorly. Moreover, existing studies only perform a binary stress assessment, while to develop a more engaging biofeedback-based application, multi-level assessment is necessary. Existing studies annotate and classify a single experience (e.g. watching a VR video) to a single stress level, which again prevents design of dynamic experiences where real-time in-game stress assessment can be utilized. In this paper, we report our findings on a new study on VR stress assessment, where three stress levels are assessed. ECG data was collected from 9 users experiencing a VR roller coaster. The VR experience was then manually labeled in 10-seconds segments to three stress levels by three raters. We then propose a novel multimodal deep fusion model utilizing spectrogram and 1D ECG that can provide a stress prediction from just a 1-second window. Experimental results demonstrate that the proposed model outperforms the classical HRV-based ML models (9% increase in accuracy) and baseline deep learning models (2.5% increase in accuracy). We also report results on the benchmark WESAD dataset to show the supremacy of the model.

✦

## 1 INTRODUCTION

THE immersive nature of Virtual Reality (VR) and the availability of devices have resulted in an influx of VR applications, going beyond the traditional realms of entertainment and gaming. For critical applications such as healthcare or simulation, another recent trend is to utilize wearable sensors to measure the physiological response of users in a VR experience. Such measurements have shown potential in different application domains as control mechanism in Autism intervention [1], and stress treatment [2].

Among physiological measurements, ECG is an attractive option due to the non-invasive nature and wide availability. Even commercial wearables such as Apple Watch can approximately calculate ECG. It has been shown in recent studies that ECG is a viable signal to measure stress [3]. The study in [4] shows the feasibility of ECG-based stress assessment in a VR environment.

However, most of these works perform binary classification of stress. In the popular WESAD dataset [3], three states are identified: neutral, stressed, and amused. However, amused is an emotional response rather than a stress-based response, so for stress assessment, the dataset reduces down to a binary classification problem (stressed vs. non-stressed). Another limitation of current studies is that typically there is a strict division of tasks and there stress labels for supervised classification. For example, in Cho et al. [4], different VR videos were shown to the users to induce mild or severe stress. However, the assumption was that during a single VR experience, the stress level will remain the same. This is not a realistic assumption, since a VR experience can be dynamic, and user's stress level may change throughout the experience.

The key application area where multi-level assessment of stress would be useful is biofeedback-based serious games. A granular level of stress assessment can provide us with the opportunity to design dynamic experiences that can change its content with the user's stress response. An example is the Bubble Bloom game for relaxation designed by our industry partner Shaftesbury Inc. and used as Task 4 in our experiments. While the game currently acts as a relaxation task only, incorporating biofeedback will help in creating a more dynamic gaming experience, where for example, the swimming speed of the fish can be controlled by a the stress level assessed through our framework. A stress level assessment *during* a VR experience/gameplay (as opposed to at the *end* of it, as typically done in current works and accompanying datasets) provides us with the opportunity to design such dynamic biofeedback-based experiences, which has widespread applications in serious games for healthcare [5], [6]. It has been shown recently that biofeedback can enhance the sense of presence in VR [7].

Furthermore, most of the current studies utilize the classical supervised learning approach of feature extraction from heart-rate variability (HRV), followed by classification. The rise of deep learning poses an interesting research

- Z. Ahmad, S. Rabbani, M.R. Zafar, and N. Khan are with the Ryerson Multimedia Research Laboratory, Ryerson University, Toronto, ON. (E-mails: z1ahmad@ryerson.ca, syeda.rabbani@ryerson.ca, muhammadrehman.zafar@ryerson.ca, n77khan@ryerson.ca).
- S. Ishaque and S. Krishnan are with the Signal Analysis and Research Laboratory, Ryerson University, Toronto, ON. (E-mails: sishaque@ryerson.ca, krishnan@ryerson.ca).

*Manuscript under review.*

question: can deep learning result in improved detection of stress from ECG? A recent study [8] reports improved stress assessment utilizing Convolutional Neural Networks (CNN). However, this study also utilizes a binary stress classification approach, and the issue of labeling a single task with a single stress label also persists.

In this paper, we aim to tackle the aforementioned issues with a new user study for stress assessment in VR, and a novel algorithm for multi-level stress assessment using CNNs. The key contributions of the work can be summarized as follows:

- We perform a user study with 9 users for multi-level stress assessment during a dynamic VR roller coaster experience. We collect multiple physiological sensor data, namely, ECG, respiration, and GSR. However, to signify the simplicity and applicability of the proposed approach, in this paper, we only utilize the ECG data to assess the level of stress [1]. This dataset will be publicly available upon publication.
- Instead of binary stress classification, we manually label the data in 10-seconds intervals from the VR roller coaster experience to three levels: low, medium, high. Due to the varying intensity in the VR experience, manual labeling results in a dataset where there is a rigorous ground truth correspondence between ECG and stress levels. Moreover, instead of simply labeling a single experience with a single label, as is done commonly, this provides us with a dynamic dataset.
- We propose a novel deep learning approach, where we fuse 1D raw ECG signal form with its spectrogram representation. Conversion to spectrogram image from 1D signals provides us with the opportunity to extract higher-level abstract features such as edges and blobs, which is not possible in 1D, as we have shown previously in the action recognition domain [9]. We further fuse the images with the raw 1D ECG signals through a novel weighted fusion scheme based on Eigenvectors. To our knowledge, this is the first approach that attempts at fusing 1D and 2D forms of ECG signal.
- We show that our proposed algorithm results in a noticeable improvement in accuracy. We compare the proposed method with traditional machine learning approaches, and ablate the results with baseline deep learning models. Besides our own dataset, we also report results on the popular WESAD dataset [3].

## 2 RELATED WORKS

### 2.1 Stress analysis from ECG

Over the past few years, numerous research has been administered using wearable devices, in order to detect/analyze physiological signals associated with stress response. Rosenberg et al. [10] observed the impact of confer-

1. Videos for the four tasks the users performed can be found at: Task 1 - baseline https://youtu.be/xW5_d5FUe5s Task 2 - roller coaster https://youtu.be/-speW58tY0M Task 3 - stroop test https://youtu.be/FmssOHjS1vc Task 4 - Bubble Bloom game https://youtu.be/E9qBZEvUCMI

ence presentation, mediation, pain, mental stress and emergency situation towards stress response, using a wireless ECG. Landolt et al. [11] conducted research which revealed that stress impaired decision making ability and reaction time of jockeys. The aforementioned research studies examine stress only through extracted ECG features, and do not utilize machine learning to classify stress.

Schmidt et al [3]. developed a dataset called WESAD, in order to detect stress and affective states through wearable devices. To our knowledge, it is the only publicly available dataset which provides correspondence between physiological signals and stress. Physiological and motion data were collected from 15 subjects and the following modalities were collected: blood volume pulse, electrocardiogram, electrodermal activity, electromyogram, respiration, body temperature, and three axis acceleration. HRV-based features were extracted for ECG, and similar classical features were extracted for other modalities (details on typical HRV-based features are provided in Section 5. Although the results utilizing classical machine learning methods such as LDA and Adaboost were encouraging, the dataset only provides means of binary classification (stress/no-stress) when it comes to assessing stress specifically, which is too simplistic for realistic applications utilizing stress as a control signal.

Recently, some studies have shown that deep learning models provide improved accuracy in stress classification compared to classical models. Song et al [12]. developed a stress classification model in order to monitor stress using deep belief networks. The model performed better than other supervised models such as SVM, RF. He et al [8]. classified acute cognitive stress from an ECG signal utilizing CNN. Compared to classical machine learning methods, they report an improved accuracy. In [13], CNN is used for photoplethysmogram (PPG) based heart rate estimation. The input to CNN is the time-frequency spectrum of the PPG-signal and tri-axial acceleration. In [14], a temporal modeling scheme based on one dimensional CNN and random forest is used to classify, detect and estimate affective states (stress and meditation) ahead of time using raw physiological and motion signals. For discussing the challenges of physiology-based affective computing in the wild, a multitask CNN is used in [15] to classify arousal, State-Trait Anxiety Inventory (STAI), stress, and valence self-reports. Authors in [16] provided study survey about the possibilities of detecting user-independent stress accurately using commercial smart watches. The effect of windowsize on recognition accuracies of affective states is also investigated in the article. However, specifically for stress assessment, most of these studies still only utilize a binary label (stress/no-stress).

Cho et al [4] is the only study that has attempted at assessing multiple levels of stress in a VR environment. They recorded photoplethysmogram (PPG), electrodermal activity (EDA), and skin temperature (SKT). Data was collected for five labels: baseline, mild stress, moderate stress, severe stress and a recovery phase. The users were shown VR videos, and were asked to perform some mental arithmetic tasks. Kernel-Based Extreme Learning Machine was able to achieve 95% accuracy using features from all three physiological signals. Since the dataset is not publicly available, we could not directly compare with their results. However, as mentioned before, only a single label was assigned to the different VR videos and tasks, which is an overly simplistic assumption for a realistic VR experience. In our study, we provide a granular level of data labeling utilizing a dynamic VR experience for a more realistic VR usage scenario.

## 2.2 Multimodal Fusion of ECG

Multimodal fusion using ECG signal has been beneficial in many applications of healthcare. In [17], different biosignals such as blood volume pulse (BVP), wrist-worn acceleration signals, the ECG signal, electromyography(EMG signal) and respiration are fed to multichannel CNN to improve the classification and accuracy of different states of mind. In [18], a deep learning based fusion framework is presented to process multimodal-multisensory bio-signals. The late fusion scheme is used in deep architecture fusion and model is evaluated by cross validation. The classification results obtained are used to explain the fact that the chest modalities contribute more to the correct classification than the wrist modalities. In [19], decision level Fusion of electroencephalography (EEG) and functional near infrared spectroscopy (fNIRS) is proposed. The fusion was based on fusing two SVM classifier decisions based on their ROC curves. The proposed fusion improves the classification accuracy in detecting stress by 10% compared to using both modalities individually. In [20], an attention-based block deep learning architecture within the device for multi-feature classification and fusion analysis is proposed. This enables the deep learning architecture to autonomously train to obtain the optimum fusion weights of different domain features. The proposed attention-based architecture has led to improving performance compared with direct connecting fusion method. To get the maximum advantage of fusing different domains, in [21], we propose a decision level fusion framework. ECG signals are converted into signal images based on R-R peaks and then signal images are made multimodal and multidomain by converting them into time-frequency and frequency domains. Finally, decision level fusion is performed on multidomain modalities. In [22], ECG heart-beats are converted into three different images using Gramian Angular Field (GAF), Recurrence Plot (RP) and Markov Transition Field (MTF) and then multimodal image fusion (MIF) is performed to create a single modality which serves as input to AlexNet. In [23], feasibility of electrogastrogram (EGG) in multi-modal mental stress assessment in conjunction with electrocardiogram (ECG) and respiratory signal (RESP) is reported. EGG, ECG, and RESP of participants were simultaneously captured and then various features were extracted from the signals, and correlation analysis was performed between mental stress levels and the features, especially the EGG features.

## 3 DATA COLLECTION AND LABELING

In this experiment [2], 9 participants were exposed to varying stress stimuli while their physiological signals was recorded. The participants' electrocardiogram (ECG), galvanic skin response (GSR) and respiration signals were measured.

2. The data collection was approved by the research ethics board of Ryerson University

An ECG sensor was used to record the electrical activity of the heart. It is a research-grade chest strap sensor provided by T-Sens with two integrated electrodes [24] (Fig. 1). Its sampling frequency was set at 256 Hz - the maximum level on the hardware being used. As seen in Fig.1 the sensor was placed around the subject's chest, while keeping the electrodes in contact with the skin.

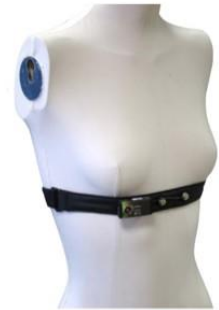

Fig. 1: ECG Sensor Placement [24]

Although we also collected GSR and respiration data, in this paper, we only attempted to assess stress levels utilizing ECG. Utilizing just ECG creates a more practical solution, since it opens up the possibilities of utilizing even commercial wearable devices such as the Apple watch. We plan to develop a multimodal assessment framework in future.

### 3.1 User Study Protocol

The videos that the participants experienced during each session are explained in detail in the next sections. The roller coaster video also contains the manual labels that we used to create the dataset.

#### 3.1.1 Preparation

Participants were first screened for cardiovascular disease, medication, and smoking. All the participants were healthy and non-smoking, and in the age range of 20-40. Participants were asked to avoid heavy exercise on the day of the study and refrain from caffeine intake an hour before. This was in order to minimize external factors that could have affected their physiological data. The experiment started off with the sensor placement and validation of the data being collected. This was followed by a short meditation session to neutralize the subject's stress level.

#### 3.1.2 Baseline

Initially we collected a baseline recording of the physiological signals under normal conditions. During this task the subjects were seated while staring at a neutral screen of floating 3D objects as seen in Fig. 2.

#### 3.1.3 Roller coaster Simulation in VR

In the second task, participants experienced a 6-minute roller coaster simulation in virtual reality. Roller coasters have been a means of inducing acute stress [25]. In this experiment we used a virtual reality simulation to recreate the real life environment of such an acute stress stimulus. The VR environment of the roller coaster is shown in Fig. 3.

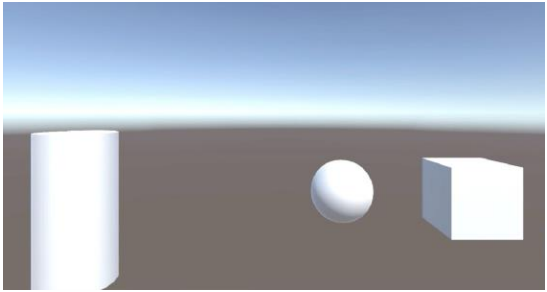

Fig. 2: Task 1 - Baseline

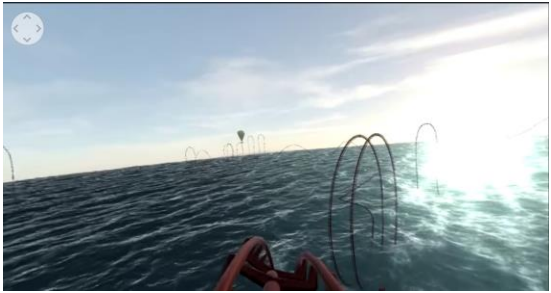

Fig. 3: Task 2 - Roller coaster Simulation in VR

#### 3.1.4 Cognitive Overload with the Stroop Test

The third task challenged the participants with the Stroop test (non-VR) - a method of inducing cognitive overload. The subjects were shown a text, the name of a colour. This text may or may not have been coloured as the name implied. They had to match the colour of the written word with three options that were displayed below the text. This task was made more challenging by setting a time limit and switching around the order of the options. An example of this is shown in Fig. 4. The Stroop test data was collected to verify whether our developed algorithm can reasonably assess stress levels in other tasks, even non-VR.

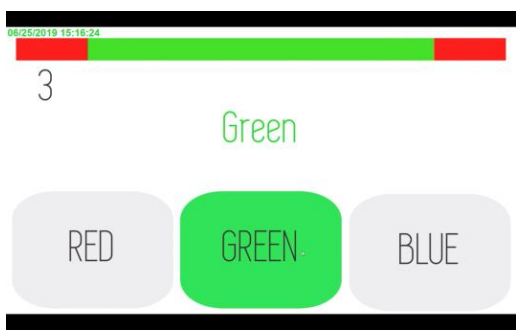

Fig. 4: Task 3 - Stroop Test

#### 3.1.5 Bubble Bloom Game in VR

The fourth task was a VR game aimed at reducing stress through its calming content. In the VR environment, the subject is underwater surrounded by a variety of vibrant fish. The subject then aims their controller and shoots bubbles to capture the fishes as shown in Fig. 5. This game was developed by our industry research partner Shaftesbury Inc., a media production company based in Toronto. Through their internal studies, Shaftesbury has validated that Bubble Bloom objectively reduces stress. We utilize this game in this task to independently validate the game's stress reduction capabilities through our model.

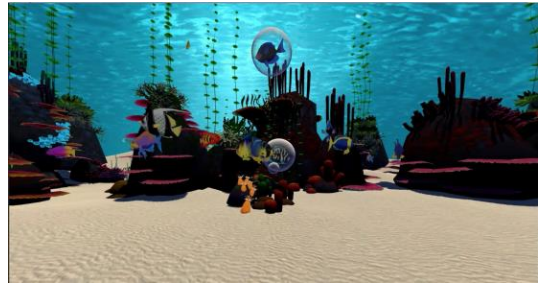

Fig. 5: Task 4 - Bubble Bloom Game

### 3.2 Obtaining Ground Truth

The data collected during the roller coaster simulation was appropriately labeled through observation of the VR experience. Typically, such tasks are rated by the participants utilizing self-reports such as Positive and Negative Affect Schedule (PANAS) [3]. However, such reports can only be collected after the experience, and it is not possible to divide the experience into smaller segments during the experience, as it will be disruptive. Our target was to rate each 10-second segment with a label so that an ML model designed for this dataset can be utilized as a control signal for such VR experiences in future. Therefore, we opted for using multiple external raters that will code the experience into small segments. Three raters were asked to label any segments of the video that they felt could induce stress. Labels 0 to 2 were assigned in order of increasing stress, based on the amount of action in the video. Segments that didn't portray a clear level of stress were left unlabeled.

| Stress Level | Example Stressor |
| --- | --- |
| 0 - **Low** | nominal conditions, no stress |
| 1 - **Medium** | moderate stress, roller coaster is in some movement with none or few loops |
| 2 - **High** | high stress, roller coaster undergoes multiple events of high activity such as continuous loops, jump scares, twists, turns and underwater dives |

TABLE 1: Examples of varying stressors in the Roller Coaster VR Simulation.

Fig.6 reflects the observations made by each rater. The inter-rater reliability can be established by Krippendorff's $\alpha$ coefficient as it accounts for many variables. In this case, the variables are unlabeled segments of the simulation, multiple raters, and ordinal data values. The acceptance level of Krippendorff's $\alpha$ is dependant on the application. When the significance of the reliability is critical, an $\alpha$ value of 0.80 or more is preferred [26]. Otherwise, an $\alpha$ value above 0.60 is also accepted respect to the use. Krippendorff's $\alpha$ calculated from the observations shown in Fig.6 is 0.44.

The $\alpha$ coefficient obtained is low possibly due to the subjective nature of the experiment. As the $\alpha$ value does not meet the acceptance standard of reliability, only segments of the data that had perfect reliability were used for training. Every 10-seconds segment of the data was evaluated, and those segments with all three raters in agreement were used in the experiments.

| Duration | C1 | C2 | C3 |
|---|---|---|---|
| 0:10 | Low | Low | Low |
| 0:20 | Low | Low | Low |
| 0:30 | Low | Low | Low |
| 0:40 | - | Medium | Medium |
| 0:50 | - | Medium | Medium |
| 1:00 | - | Medium | Medium |
| 1:10 | - | Medium | High |
| 1:20 | - | High | High |
| 1:30 | - | High | High |
| 1:40 | Medium | High | - |
| 1:50 | Medium | High | - |
| 2:00 | Medium | Medium | - |
| 2:10 | Medium | High | - |
| 2:20 | High | High | - |
| 2:30 | High | High | Medium |
| 2:40 | Medium | Low | Medium |
| 2:50 | Medium | Low | Medium |
| 3:00 | Medium | Medium | Medium |
| 3:10 | High | Medium | Medium |
| 3:20 | High | Medium | Medium |
| 3:30 | High | Medium | Medium |
| 3:40 | Medium | Medium | Medium |
| 3:50 | Medium | Medium | Medium |
| 4:00 | Medium | Medium | Medium |
| 4:10 | Medium | High | High |
| 4:20 | High | High | High |
| 4:30 | High | High | High |
| 4:40 | High | High | High |
| 4:50 | High | High | High |
| 5:00 | High | High | High |
| 5:10 | High | High | High |
| 5:20 | High | High | High |
| 5:30 | High | High | High |
| 5:40 | High | High | High |
| 5:50 | High | High | High |
| 6:00 | High | High | High |

Fig. 6: Raters' labels

Such manual labeling solely focuses on the VR experience in hand to label the level of stress, providing a true correspondence between physiological signals and stress. Due to the dynamic nature of the roller coaster experience, it was possible to assign different levels of stress to different segments based on the intensity of the experience itself. The roller coaster video, with the manually annotated labels can be seen at https://youtu.be/-speW58tY0M .

This collected dataset will be called the **Ryerson Multimedia Research Laboratory (RML)** dataset for the rest of the paper.

## 4 PROPOSED FUSION METHOD

In the proposed fusion method, we use ECG data in raw form i.e in one-dimensional (1D) form and in two-dimensional (2D) form by transforming raw ECG data into spectrograms. We trained a 1D Convolutional Neural Network (CNN) on raw ECG data and ResNet-18, a CNN model [27], on spectrograms. Finally, learned features are extracted from both CNNs, and we perform weighted average fusion of both modalities, 1D form of ECG and spectrograms. Fused features are used as input to a Support Vector Machine (SVM) classifier for improving the classification accuracy. The SVM classifier is trained separately from the CNNs. We have previously shown using CNNs as feature extractors followed by SVM as a classifier results in a slight increase in accuracy for similar fusion architectures compared to using CNN (softmax) for classification directly [28]. The complete overview of the proposed fusion method is shown in Fig. 7. The weighted average fusion network is shown in Fig. 9.

### 4.1 Raw ECG data with 1D CNN

When used as 1D representation, we used one-second (1s) window to generate 1D snippets from raw ECG data. We trained ID CNN on these snippets. The input to ID CNN is a snippet of 1 x 256. Five kernels of size 1 x 5 are used in first convolutional layer, followed by 1 x 2 subsampling layer. The second convolutional layer contains 10 filters of same size followed by 1 x 2 subsampling layer. The third convolutional layer contains 10 filters of size 1 x 4 followed by 1 x 2 subsampling layer, a fully connected layer and a classification layer.

### 4.2 ECG to Spectrogram Transformation

For using as 2D representation, we generate spectrograms from one-second (1s) ECG segments using short-time Fourier transform (STFT). The formation of spectrogram from ECG signal is based on the Short-Term Fourier Transform (STFT) below:

$$S_i(k, l) = \sum_{m=0}^{M-1} x_i(m) w(l-m) e^{-j \frac{2\pi}{M} km} \quad (1)$$

where $w(.)$ is the window function, e.g., Hanning window and $S_i(k, l)$ is the STFT of ECG segment $x_i$. The spectrogram is then calculated as the square modulus i.e. $|S_i(k, l)|^2$.

As examples, spectrograms of three levels of stress for RML data are shown in Fig. 8.

We used ResNet-18 to train on spectrograms. For performing weighted average fusion, learned features are extracted from fully connected layer of 1D CNN and from "pool5" layer of ResNet-18 as shown in Fig. 7. We extracted features from "pool5" layer of ResNet because this is the second last of ResNet and it has more learned features than other layers [29].

### 4.3 Weighted Average Fusion

Let $F_1$ and $F_2$ are the learned features extracted from both the modalities as shown in Figures 7 and 9. The steps followed for weighted average fusion performed in Fig. 9 are as below:

1) We calculate the covariance matrix for both features $F_1$ and $F_2$ using the following equations.

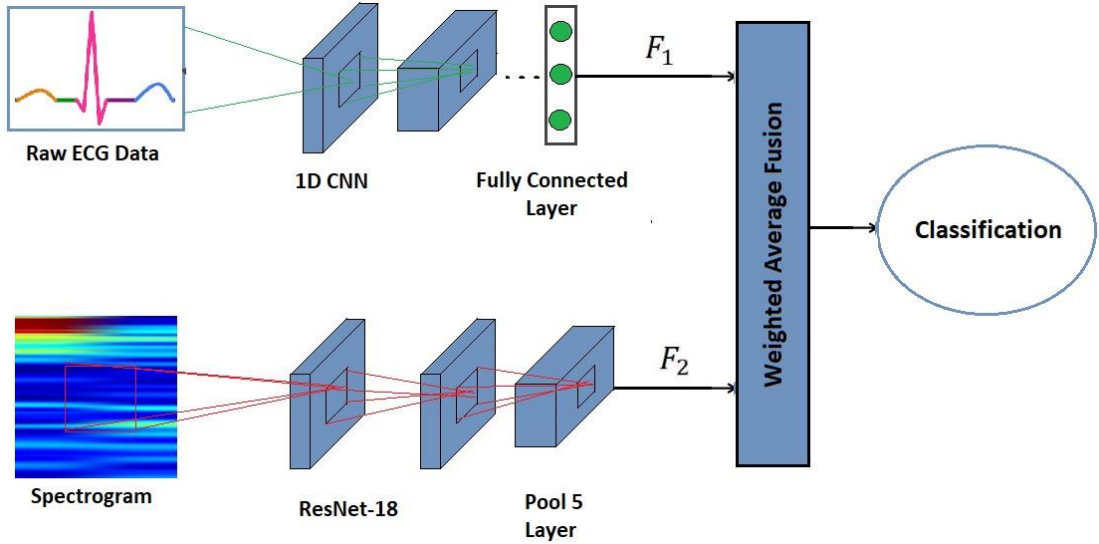

Fig. 7: Complete Overview of the Proposed Fusion Method.

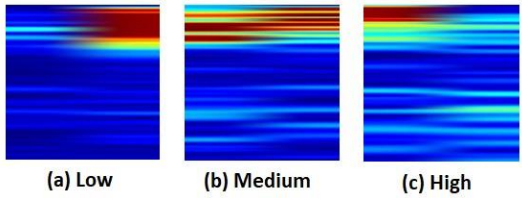

Fig. 8: Spectrograms of three levels of stress for RML data

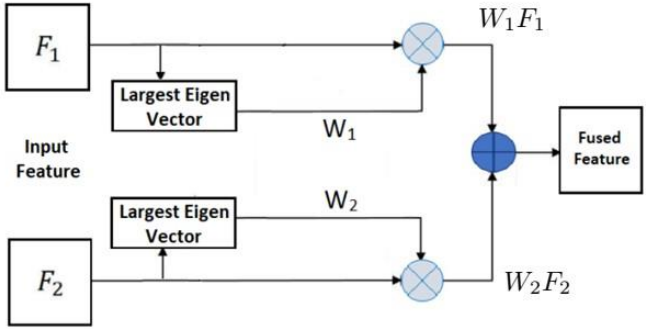

Fig. 9: Weighted Average Fusion Network.

$Cov(F_1) = F_1^T F_1$
$Cov(F_2) = F_2^T F_2$

2) We then compute the eigenvectors $V_1$ and $V_2$ and eigenvalues $d_1$ and $d_2$ from both covariance matrices $Cov(F_1)$ and $Cov(F_2)$ respectively and then sorted them in descending order.
3) Extract the eigen vectors from the first columns of $V_1$ and $V_2$. These eigen vectors correspond to the largest eigen values and are considered as weighted vectors $W_1$ and $W_2$ for features $F_1$ and $F_2$ respectively.
4) We column-wise multiply weighted vector $W_1$ with the feature matrix $F_1$ and $W_2$ with the feature matrix $F_2$ and finally add these products to get the final fused features. If $F$ is the fused feature matrix then mathematically it is written as

$$F = W_1 F_1 + W_2 F_2 \qquad (2)$$

Principal Component Analysis serves as the inspiration behind our weighted fusion scheme. By utilizing the eigenvectors corresponding to the largest eigenvalues as weight coefficients, we ensure enhancement of the discriminative components among the features.

## 5 CLASSICAL FEATURES AND ML MODELS

The classical machine learning models were trained on features extracted from the participants' ECG signal. The duration between successive R peaks is usually used to compute time and frequency domain features that analyse a person's HRV. HRV features have been found to be highly correlated with stress and hence are used widely in stress detection studies that employ classical machine learning [30]. The following features were utilized for the classical models:

### 5.1 Time Domain Features

Time domain features are the basic statistical features extracted from ECG signal data. We extracted the following statistical features, following the work in [30]:

1) **Heart Rate (HR):** Heart rate is the number of beats or R peaks that take place in one minute.
2) **Root Mean Square of Successive Differences (RMSSD)**: RMSSD is often utilized to analyze short term HRV, but it is also possible to analyze it through ultra-short-term periods, which ranges from 0-2 minutes [31].
3) **Average Value of N-N Intervals (AVNN)**: AVNN is the average of all the R-R intervals (also known as N-N intervals).
4) **Standard Deviation of N-N Intervals (SDNN):** The standard deviation of NN interval measures the total power with respect to the segment being analyzed, it also represents the total variability associated with HRV and it is measured in ms [31]. Short term 5 minute windows are often utilized to analyze SDNN, but it can also be efficiently measured using ultra-short windows.

5) **pNN50:** Represents the percentage difference associated with R-R interval which differ more than 50 ms. It can be effectively analyzed within 2-minutes and some researchers used it to analyze signals from an ultra-short window of 60 seconds [31]. It is proportional to HRV, increases with increasing HRV and decreases during increased levels of stress.

### 5.2 Frequency Domain Features

The frequency domain features were calculated by computing the Power Spectral Density (PSD) of the ECG signals using the Lomb-Scargle method. The Lomb-Scargle method is best known for handling uneven samples such as the R-R intervals in this data [32].

1) **Very Low Frequency (VLF):** The VLF band in frequency domain ranges between 0.0033-0.04 Hz. VLF is predominantly affected through HRV associated with thermoregulatory and physical acitivity. VLF amplitude and frequency are stimulated by increase in stress only, since it is regulated by SNS activity.
2) **Low Frequency (LF):** The LF band ranges between 0.04-0.15 Hz within the PSD. Stressful events which are correlated to increased HR and SNS activity, result in an increase in LF values.
3) **High Frquency (HF):** The HF band ranges between 0.15-0.40 Hz within the PSD. Precise HF values can be obtained from ultra-short-term period of 1 minute [31]. It shares a high correlation with HRV, increasing or decreasing along with HRV. It also shares a strong correlation with RMSSD and pNN50 from time domain. High HF occurs as a result of high HRV,
4) **Total Power Spectrum (TP):** The total power spectrum is the sum of all NN intervals up to a frequency of 0.04Hz [30].

### 5.3 Classical Machine Learning Models

A serious limitation of traditional supervised ML is, it requires an extensive domain knowledge to find an optimal model and ideal parameters to solve a problem [33]. This limitation has stimulated the research interest in automatic ML (aML). aML is a new paradigm in ML where, the target is to develop an approach that automatically finds an optimum model and solve the problem without human interaction [34]. The popular aML frameworks are Auto-sklearn [34], H2O.ai [35], Auto-WEKA [36], Recipe [37], Autostacker [38] and Tree-based Pipeline Optimization Tool (TPOT) [39]. The primary focus of these frameworks is hyper-parameters tuning and optimal model selection.

We first show that ad-hoc tuning of popular classical models results in sub-optimal accuracy. Then, we utilized aML to pick the best model for us automatically. Unlike other frameworks, the TPOT aML framework also provides the functionality of feature engineering and feature selection by evaluating pipelines based on cross validation scores. Therefore, we used TPOT to conduct our experiments. TPOT uses a genetic algorithm to create an initial population of various ML models and on each iteration to perform a fitness test. That is a natural way to keep only the best performing models for the next iteration. It allows us to test a large range of various ML models, and build a model on the top of best performing generation, that makes it more efficient as compared to ad-hoc approaches.

## 6 EXPERIMENTAL RESULTS

### 6.1 Datasets

We experiment with the RML dataset presented before, and the WESAD dataset [3]. WESAD is a multimodal dataset recorded with a chest-worn sensor recording ECG, accelerometer, EMG, respiration, and body temperature and a wrist-worn sensor recording PPG, accelerometer, electrodermal activity, and body temperature. 9 participants were included in data collection activities, each for approximately 100 minutes. Initial experiments on this dataset were carried out for detecting and differentiating three affective states (neutral, stress, amusement). For our experiments, we only used raw data obtained from the chest-worn ECG Signal.

Since ECG has high inter-subject variablity, we utilize leave-one-subject-out (LOSO) cross-validation (CV) for all our experiments, where the models are trained with data from all-but-one subjects, and tested on the held out subject. The average results across all subjects are presented.

### 6.2 Classical Machine Learning

To align with the 1-second window of our deep learning model, we use the shortest possible window of 4 seconds for the classical models. Anything below 4 seconds resulted in detrimental performance due to the statistical nature of the features. For model search using TPOT, we divided the datasets into two subsets, training and testing by randomly splitting 85% data into training and 15% data into testing samples. Later we used the parameters of optimized model returned by TPOT to train the model and validate it using LOSO CV.

First, we trained individual ML classifiers i.e. Random Forest (RF), K-Nearest Neighbours (KNN), Support Vector Machines (SVM) and XGBoost Classifier (XGB) from the scikit-learn package with default parameters and LOSO cross validation on the RML dataset. We also utilize a bagging classifier which is a meta-estimator ensemble approach. RF was used as a base model in bagging classifier with 100 estimators.

The poor performance of classical ML models motivated us to use the aML framework TPOT for hyper-parameters tuning and optimal model selection. The genetic programming parameters for TPOT were set as: generation=400, population=100 and cv=10. Table 2 and Table 3 shows the results of individual ML classifiers and optimized model using TPOT.

### 6.3 Proposed Fusion Method

Similar to the classical ML approach, we used leave-one-subject-out (LOSO) cross-validation (CV) setting on both datasets to train and test our proposed fusion model. All the hyperparameters for the proposed model were tuned using a grid search approach on the training set.

| Classifier | Accuracy | Precision | Recall | F1-Score |
|---|---|---|---|---|
| Bagging Classifier | 53.9 | 23.9 | 31.2 | 27.1 |
| TPOT - KNN | **57.1** | 29.0 | 30.5 | 27.4 |
| RF | 46.2 | 33.5 | 33.4 | 30.4 |
| XGB | 52.1 | 32.6 | 31.5 | 32.0 |
| SVM | 46.2 | 19.0 | 30.1 | 23.3 |
| KNN | 45.1 | **34.1** | **33.9** | **33.9** |

TABLE 2: Performance of classical ML models on the RML dataset. Best method in bold.

| Classifier | Accuracy | Precision | Recall | F1-Score |
|---|---|---|---|---|
| Bagging Classifier | 64.6 | 52.1 | 50.0 | **51.0** |
| TPOT - XGB | **66.2** | 49.4 | 50.8 | 50.1 |
| RF | 61.8 | **52.5** | 49.3 | 50.8 |
| XGB | 65.8 | 48.0 | **51.0** | 49.5 |
| SVM | 60.2 | 47.3 | 46.0 | 46.6 |
| KNN | 57.2 | 49.2 | 50.0 | 46.9 |

TABLE 3: Performance of classical ML models on the WESAD dataset. Best method in bold.

To validate the significance of the proposed weighted average fusion method, we also perform ablation experiments with average fusion. In average fusion, the the weight vectors $W_1$ and $W_2$ are replaced with the scalar $0.5$ in Equation 2. The experimental results on the RML and the WESAD dataset are shown in Table 4 and Table 5 respectively. Comparison of accuracy and $F_1$ score of proposed fusion method on WESAD dataset with previous methods for the three-class Problem (neutral, stress, amusement) using LOSO and chest-worn ECG is shown in Table 6.

| Methods | Accuracy | Precision | Recall | F1 Score |
|---|---|---|---|---|
| Raw ECG with 1D CNN | 62.8 | 62.5 | 62.8 | 61.6 |
| Spectrograms with ResNet-18 | 63.4 | 64.3 | 64.9 | 62.3 |
| Average Fusion | 64.1 | 65.2 | 64.1 | 63.3 |
| Proposed Fusion Method | **66.6** | **67.6** | **66.6** | **65.6** |

TABLE 4: Performance of deep learning models on the RML dataset. Best method in bold.

| Methods | Accuracy | Precision | Recall | F1 Score |
|---|---|---|---|---|
| Raw ECG with 1D CNN | 51.4 | 51.2 | 51.4 | 48.7 |
| Spectrograms with ResNet-18 | 72 | 75.1 | 72 | 72.5 |
| Average Fusion | 69 | 74.2 | 69 | 69.1 |
| Proposed Fusion Method | **72.7** | **76.6** | **72.7** | **73.1** |

TABLE 5: Performance of deep learning models on the WESAD dataset. Best method in bold.

| Previous Methods | Accuracy | F1 Score |
|---|---|---|
| P.Schmidt et al. [3] | 66.29 | 56.03 |
| **Proposed Fusion Method** | **72.7** | **73.1** |

TABLE 6: Comparison of performance of the proposed fusion Method on the WESAD dataset with previous methods.

## 7 DISCUSSION

We first discuss the results from the classical ML models. From able 2 and Table 3 we can see that all the classical ML models, including aML, results in poor performance, topping off only at 57.1% and 66.2% accuracy, respectively, which were achieved through the aML models. However, we see that while aML resulted in an increase in accuracy - precision, recall and F1-score takes a hit, where the ad-hoc tuned classical ML models achieve better results. This is not surprising, since aML is typically specifically geared towards accuracy improvement, which is the main performance metric used in ML literature. We also see that TPOT picked KNN as the optimal classifier for the RML dataset, and XGB for the WESAD dataset. This is also expected, since aML searches through multiple models to pick the optimal classification strategy.

Looking at the results for the proposed fusion model in Table 4 and Table 5, we can see that for the RML dataset, even the baseline 1D CNN model outperforms the classical ML models. For WESAD, the 1D CNN model performs poorly, but the spectogram-based model provides a significant boost, showing the power of CNNs for 2D signals. Finally, we see that our proposed weighted fusion model outperforms all the other models, including the average fusion. The performance imporovement holds across all the metrics. An intetesting obesrvation is that the baseline average fusion actually results in a decrease in accuracy for the WESAD dataset when compared to the spectrogram model (Table 5 rows 2 and 3). This shows that blindly fusing the available features may not always result in a robust model. However, our weighted fusion approach overcomes this limitation by intelligently fusing the discriminative components of the features.

Finally, in Table 6 we show that our proposed model significantly outperforms the state-of-the-art on the 3-class classification problem from the WESAD dataset. It is important to note that although there are other results reported on WESAD in literature, we found that only [3] reported results from solely the ECG signal and utilizing LOSO cross validation. There are two other recent works that report results on ECG only [18], [40]. However, they do not perform LOSO cross validation, which drastically reduces the generalizability, since the models have access to data from all subjects during training. Our own previous method [21] fails to generalize on the data when LOSO cross-validation is employed. However, the proposed weighted fusion of 1D signal and spectrogram can successfully generalize as we can see from the results.

### 7.1 Performance of the trained model on other tasks

We also validate the trained model by classifying the stress levels of the other three tasks during the user experiments, namely, baseline task, Stroop test, and Bubble Bloom game. No training was performed from the data collected through these tasks. The purpose of this experiment is to qualitatively assess the performance of our trained model on different tasks.

During the baseline task, we observed a constant stress level of "low" for almost all users. This goes to show that the deep learning model is successful in classifying the baseline stress. Similarly, for the Bubble Bloom game, the stress levels remained between low and medium, verifying the calming content of the game which consistently resulted in low stress. Finally, for the Stroop test, high stress levels were observed for the participants. This qualitative observations verify that the trained deep learning model can be utilized in dynamic VR applications for real-time assessment of stress. Since these tasks were not annotated, we do not report any quantitative results.

## 8 CONCLUSION

In this paper, we report findings on a stress assessment study in a VR environment utilizing ECG signal. We collect data on 9 users experiencing a VR roller coaster experience. Unlike existing approaches which label a single experience with a single binary label (stress/no-stress), we perform a more granular labeling by categorizing different parts of the dynamic VR experience into tree levels of stress. We propose a novel multimodal deep fusion approach utilizing spectrograms and raw 1D ECG signals that outperforms classical ML models and baseline deep learning models. We also report results on the benchmark WESAD dataset, where our proposed model outperforms the existing ECG-based approaches. Utilizing a single modality makes the proposed approach attractive for commercial applications that can utilize consumer devices such as the Apple watch.

While the multi-level assessment has potential to be applied in bio-feedback applications, the efficacy of such approach remains to be seen. In future, we plan to incorporate the framework into serious games to determine the efficacy. As mentioned before, the Bubble Bloom game utilized in Task 4 will be an ideal candidate, where the game content (e.g. speed of fish swimming) can be adapted based on the stress level of the user for a dynamic experience.

The collected dataset also has room for improvement. Currently we have data for only 9 users, which can potentially impact robustness. Moreover, the labeling process utilizes only 3 raters, where involvement of more raters (perhaps via crowd-sourcing) will create a more diverse ground truth set.

## ACKNOWLEDGMENTS

NSERC's financial support for this research work through a Collaborative Research and Development (CRD) Grant (# 537987-18) is much appreciated.